\documentclass[10pt,twocolumn,letterpaper]{article}

\usepackage[datasets]{wacv}

\usepackage{graphicx}
\usepackage{booktabs}
\usepackage{tabularx}
\usepackage{array}

\definecolor{wacvblue}{rgb}{0.21,0.49,0.74}
\usepackage[pagebackref,breaklinks,colorlinks,allcolors=wacvblue]{hyperref}
\hypersetup{
  pdftitle={ARCH-B: Architectural Representation, Comprehension and Hierarchy Benchmark},
  pdfauthor={Kieran Sagar Parikh; Jose Luis Garcia del Castillo y Lopez}
}

\title{ARCH-B: Architectural Representation, Comprehension and Hierarchy Benchmark}
\author{
  Kieran Sagar Parikh\\
  Harvard University\\
  {\small kieran\_parikh@mde.harvard.edu}
  \and
  Jose Luis García del Castillo y López\\
  Northeastern University\thanks{Work partially conducted while at Harvard University.}\\
  {\small garciadelcastillo@northeastern.edu}
}

\begin{document}
\maketitle

\begin{abstract}
Multimodal models increasingly interpret visual environments, but their ability to recognize the same building across photographs, floor plans, elevations, sections, and renderings remains poorly characterized. We introduce ARCH-B, a benchmark of 354 four-choice questions across 11 cross-representational archetypes, constructed from a building-linked corpus of 3.9 million architectural images using visually similar distractors, model-guided difficulty screening, and manual validation. We evaluate 25 multimodal models and collect 5,830 responses from non-expert human participants. Model accuracy ranges from 10.45\% to 83.90\%, compared with a human baseline of 35.35\%. Models perform comparatively well on mixed-representation outlier detection and photograph matching, but remain weaker on floorplan-to-photograph correspondence. Human and model difficulty across archetypes is only weakly correlated (Spearman's \(\rho=0.33\)). Held-out evaluation confirms that the difficulty identified during screening generalizes beyond the curation models. ARCH-B provides a diagnostic evaluation of visual correspondence and representation transfer across architectural media.
\end{abstract}

\section{Introduction}
\label{sec:intro}

Architectural visual understanding is inherently cross-representational. The same building may appear as a photograph, a floor plan, an elevation, a section, a project rendering, a small set of partial views or a combination thereof that must be recognized as belonging to the same object. For humans, these transitions are routine. Architects move constantly between abstract drawings and embodied views of space, and non-architect occupants do something similar whenever they use maps, plans, signage, or partial visual cues to orient themselves in the built environment. For multimodal large language models, however, this family of abilities remains difficult to evaluate systematically; existing multimodal benchmarks probe expert reasoning, core visual perception, and 2D and 3D spatial understanding \citep{yue2024mmmu,fu2024blink,tong2024cambrian,yang2025thinking,ma2025threedsrbench}, but do not directly test whether models can preserve building identity across photographs and architectural drawings.

We present ARCH-B, a multimodal benchmark of cross-representational architectural comprehension. We define \emph{cross-representational and hierarchical architectural comprehension} as the selection and alignment of visual evidence for building identity across changes in viewpoint, abstraction, and representational convention. The hierarchy in ARCH-B refers to movement across levels of architectural representation, from local perspectival views of interiors and exteriors to whole-building drawings such as elevations, sections, and floor plans.

ARCH-B consists of 354 multiple-choice question samples, each belonging to one of 11 question archetypes. The 11 question archetypes cover image-drawing correspondence, floor-plan matching, mixed-representation outlier detection, interior-exterior correspondence, and same-building recognition across photographs and drawings. Recent architectural benchmarks emphasize floor-plan semantics, drawing literacy, higher-level spatial cognition, or principle-grounded engineering reasoning \citep{ganon2025waffle,kondratenko2026aecvbench,shen2026archsibench,du2026mmarch}; ARCH-B instead isolates correspondence and identity across heterogeneous representations of the same building. Our premise is that an architectural benchmark should test not only whether a model can describe a scene, but whether it can align representations that differ in viewpoint, abstraction, and representational conventions.

The images used for the questions were sourced from a large architectural image corpus that we assembled from real-estate listings and architectural project pages. Question samples were generated programmatically from that corpus, using different algorithms for each question archetype. Samples were selected for inclusion in the final benchmark through LLM-in-the-loop difficulty selection against four strong multimodal models, followed by manual validation and balancing.

We used the benchmark to evaluate 25 multimodal models, with Gemini 3.1 Pro Preview achieving a top score of 83.90\%. The leaderboard spans a wide range of performance, and the strongest proprietary models substantially outperform both chance and the non-expert human baseline. We also conducted a human study through a human-subjects crowdsourcing platform, collecting 5,830 valid responses across the 354 benchmark questions. The response-weighted human baseline was 35.35\%: above random chance, but far below the best evaluated models.

The main contributions of this paper are threefold. First, we introduce and publicly release ARCH-B, a 354-question benchmark for evaluating cross-representational architectural comprehension across photographs, floor plans, elevations, sections, and renderings. Second, we describe a scalable generation and curation pipeline that combines a large architectural image corpus, embedding-based candidate construction, LLM-in-the-loop difficulty selection, and manual validation by researchers. Third, we provide a broad empirical evaluation of 25 multimodal models alongside a non-expert human baseline, showing that current frontier models can perform strongly on these tasks while substantial variation remains across archetypes and models.

\section{Related Work}
\label{sec:related-work}

\subsection{Multimodal visual and spatial reasoning benchmarks}
\label{sec:multimodal-benchmarks}

Visual question answering benchmarks have progressed from object recognition and language-conditioned prediction toward compositional and expert-level reasoning. GQA uses scene graphs and functional programs to test structured reasoning over real images \citep{hudson2019gqa}, while MMMU evaluates models using diagrams, charts, maps, and other expert materials across multiple disciplines \citep{yue2024mmmu}. Because such broad evaluations can conflate perception, knowledge, and reasoning, vision-centric benchmarks isolate perceptual capabilities more directly. MMVP exposes visual distinctions poorly represented by common vision--language encoders \citep{tong2024eyes}; BLINK reformulates classic computer-vision tasks, including visual correspondence, relative depth, and multi-view reasoning, as multiple-choice evaluations \citep{fu2024blink}; and CV-Bench tests 2D relations and counting alongside 3D depth and distance judgments \citep{tong2024cambrian}.

Recent benchmarks extend spatial evaluation across views and over time. VSI-Bench tests whether models can recover and query a representation of an indoor environment from video \citep{yang2025thinking}, while 3DSRBench evaluates object-centered 3D relations and robustness to unusual viewpoints \citep{ma2025threedsrbench}. ARCH-B addresses a complementary problem: the identity that must persist is an entire building, and observations vary in both viewpoint and representational convention. Matching photographs, floor plans, elevations, sections, and renderings requires transferring evidence between projective images and abstract drawings rather than reasoning within a single visual formalism.

\subsection{Architectural imagery, drawings, and multimodal evaluation}
\label{sec:architectural-evaluation}

Architectural vision datasets have largely supported specialized prediction tasks. CubiCasa5K, CubiGraph5K, and FloorPlanCAD provide raster, graph, and vector floor-plan representations for parsing and symbol recognition \citep{kalervo2019cubicasa5k,lu2021organizational,fan2021floorplancad}. ZInD aligns panoramas, room layouts, and 2D and 3D floor plans to support layout estimation and multi-view registration \citep{cruz2021zind}, while WAFFLE collects nearly 20,000 in-the-wild floor plans and associated metadata across diverse building types \citep{ganon2025waffle}. These resources advance parsing, layout recovery, and geometric registration, but do not evaluate identity across heterogeneous architectural representations.

Recent benchmarks also evaluate general-purpose multimodal models on architecture and engineering. AECV-Bench tests counting, OCR, and drawing-grounded question answering on floor plans and engineering drawings \citep{kondratenko2026aecvbench}. ArchSIBench evaluates architectural spatial intelligence through tasks involving perception, navigation, transformation, configuration, circulation, and functional zoning \citep{shen2026archsibench}. MMArch instead requires models to combine technical figures with architecture and civil-engineering knowledge \citep{du2026mmarch}. These benchmarks emphasize drawing literacy, broader spatial cognition, or professional knowledge; ARCH-B focuses specifically on recognizing multiple visual representations as evidence of the same built object.

\subsection{Benchmark construction and difficulty control}
\label{sec:difficulty-control}

Automatically generated benchmarks provide scale but risk artifacts, trivial distractors, and misaligned difficulty. GQA uses programmatic generation from scene graphs \citep{hudson2019gqa}; adversarial filtering and AFLite use model behavior to reduce exploitable biases \citep{lebras2020aflite}; and MMMU-Pro removes questions answerable without images and expands answer sets to reduce shortcuts \citep{yue2024mmmupro}. ARCH-B combines structured generation with visually similar distractors and model-guided difficulty filtering for building-level correspondence. Manual validation and held-out model evaluation further address item validity and test whether the selected difficulty generalizes.

\section{Methods}
\label{sec:methods}

We constructed ARCH-B by assembling an architectural image corpus, generating candidate questions, filtering them with four multimodal models, and manually validating the selected questions.

\subsection{Corpus of Architectural Imagery}
\label{sec:corpus}

We built the source corpus by scraping architectural imagery from two source families: real-estate listings and architectural project pages. Real-estate listings were sourced from Zillow and Realtor.com, while architectural projects were sourced from ArchDaily and Deezeen. We scraped daily from November 2024 until October 2025, producing a corpus of 3.9 million images, including 58,000 floor plans, drawn from 95,000 real-estate listings and 41,000 architectural projects. Crucially, we chose to preserve relationships among images and their parent listing or project, rather than flattening the corpus into isolated files, enabling us to query images by listing or project during programmatic question sample generation. We refer to this grouping of images and associated metadata as \emph{building record}.

We augmented this corpus with vector embeddings that would be needed for programmatic question sample generation, detailed in \cref{sec:generation}. We encoded each image using OpenCLIP (an open-source implementation of CLIP that learns a shared representation space for images and natural-language text) using the ViT-B/32 architecture and the laion2b\_s34b\_b79k pretrained checkpoint, producing a 512-dimensional embedding for each image \citep{radford2021clip,cherti2023openclip,schuhmann2022laion5b}. These embeddings can be used to identify visually related images using the cosine similarity between two images’ embeddings, as well as to perform zero shot image classification by embedding a label string using the same OpenCLIP encoder and measuring the cosine similarity of the image embedding and the label string embedding.

\subsection{Question Sample Generation}
\label{sec:generation}

Every benchmark question sample follows the same format: a short natural-language instruction, optionally one or more reference images in the question stem, and four answer choices with exactly one correct answer. Answer choices do not include descriptive text beyond an answer ID. All benchmark questions use a four-alternative forced-choice format with exactly one correct answer, so chance performance is 25\%.

\begin{figure}[t]
  \centering
  \includegraphics[width=\columnwidth]{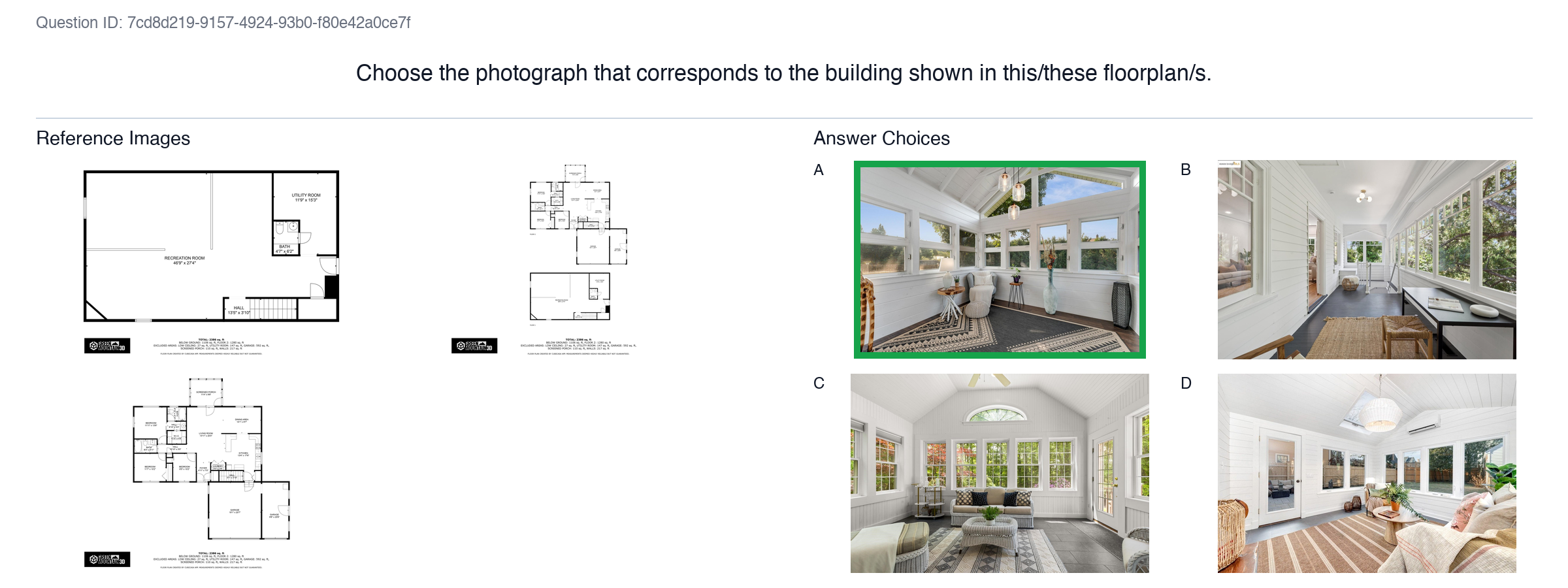}
  \caption{Example question for archetype 2. The correct answer is outlined in green.}
  \label{fig:example-question}
\end{figure}

Each question sample belongs to one of 11 question archetype. The question archetypes were designed by the authors to sample all possible cross-representational and outlier permutations of representations and hierarchy types. The archetype to which a question belongs defines its natural-language instruction and the types of images that appear as reference images and answer choices. For example, question samples belonging to archetype 1 will all have the natural-language instruction “Choose the photograph that corresponds to the building shown in this/these floorplan/s,” and will have floorplans as reference images and photographs as answer choices (\cref{fig:example-question}). A complete list of archetype metadata, including a representative question sample for each archetype, is provided in \cref{app:archetypes}.

Given that each question archetype can be defined by a set of rules, we designed generation algorithms to generate question samples for each archetype. This approach follows the broader use of structured annotations and rule-based generation to construct diagnostic visual-reasoning questions \citep{hudson2019gqa}. Each archetype has its own generation algorithm, but all generation algorithms share the same shape. Generation begins by querying the corpus for building records that satisfy the image-type requirements of a target archetype. Image types (floorplan, interior, exterior, etc) are identified using zero shot image classification on the multimodal OpenCLIP embeddings described in \cref{sec:corpus}. For example, a question sample for archetype 1 (“match photograph to floorplan”) can only be created from a building record that contains at least one floor plan and at least one photograph, while a mixed-drawing question sample (archetype 11) requires a building record that contains a floor plan, an elevation, a section, and at least one photograph. Once a building record has been selected, question images and the correct answer choice are sampled randomly from its eligible images, as determined by the precomputed image-type labels. Incorrect answer choices are sampled randomly from all eligible images of the required image type across the corpus.

However, after an initial round of question sample generation using random image selection , we observed that many generated question samples were either too easy or were unsolvable. We therefore introduced two embedding-based methods to improve image selection during question sample generation: diverse image selection within a property using agglomerative clustering, and distractor (incorrect answer choice) selection via visual similarity.

\begin{figure}[t]
  \centering
  \includegraphics[width=\linewidth]{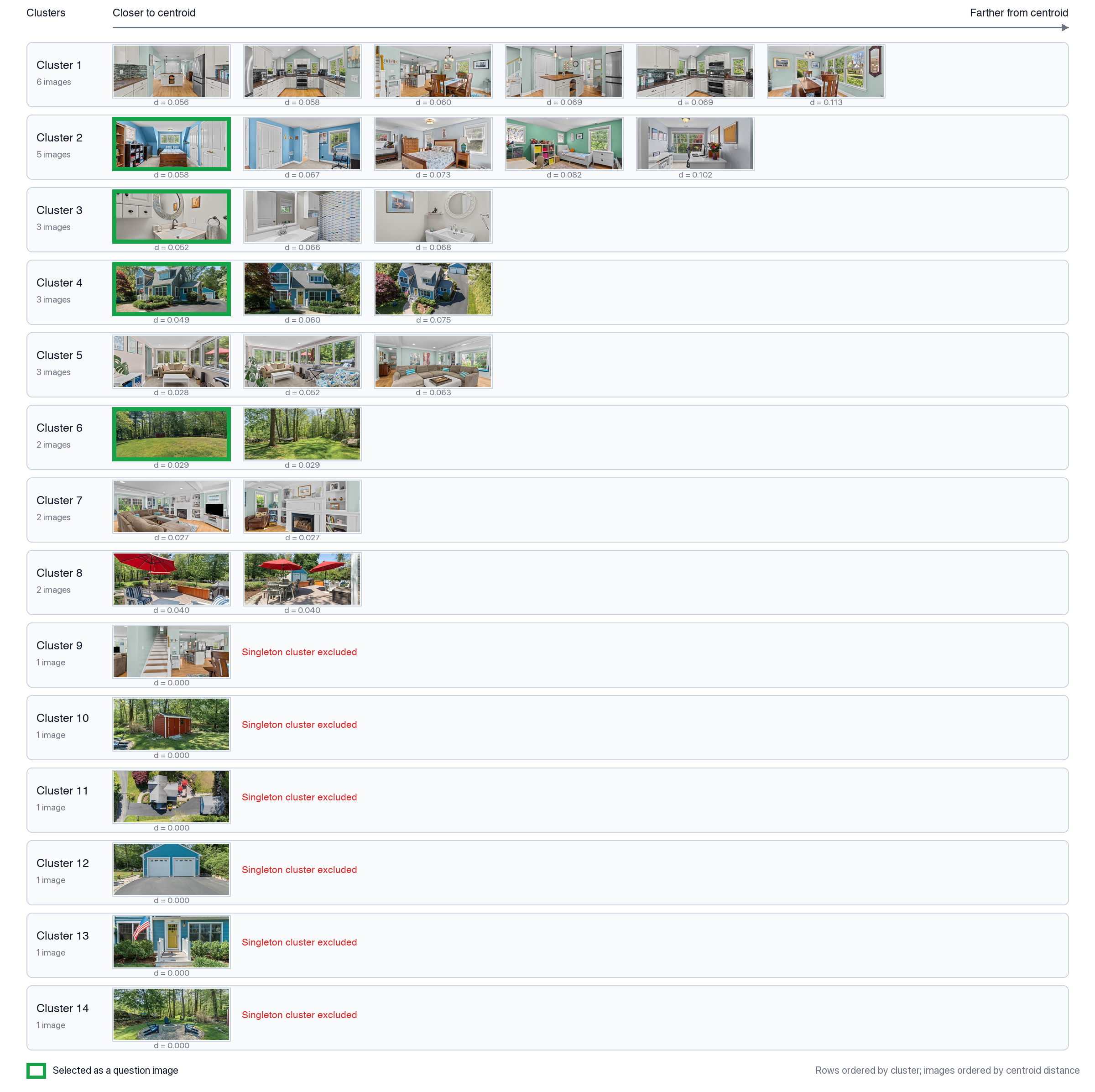}
  \caption{Cluster-based reference image selection, promoting both representativeness and visual diversity; selected images are outlined in green.}
  \label{fig:cluster-selection}
\end{figure}

When selecting multiple representative images of a building record, we found that random selection would sometimes produce a set of images that were similar to each other, limiting the visual information about the building as a whole. We therefore developed a method to perform diverse image selection within a property using agglomerative clustering. We cluster candidate images with a distance threshold of 0.25, discard singleton clusters, and select centroid-nearest representatives that maximize visual diversity. This makes the reference set less redundant and forces the model to compare distinct views of the same property rather than nearly duplicated frames. If too few multi-image clusters survive, we fall back to round-robin selection across available clusters, starting with centroid-nearest images and then sampling additional images as needed.

\begin{figure}[t]
  \centering
  \includegraphics[width=\columnwidth]{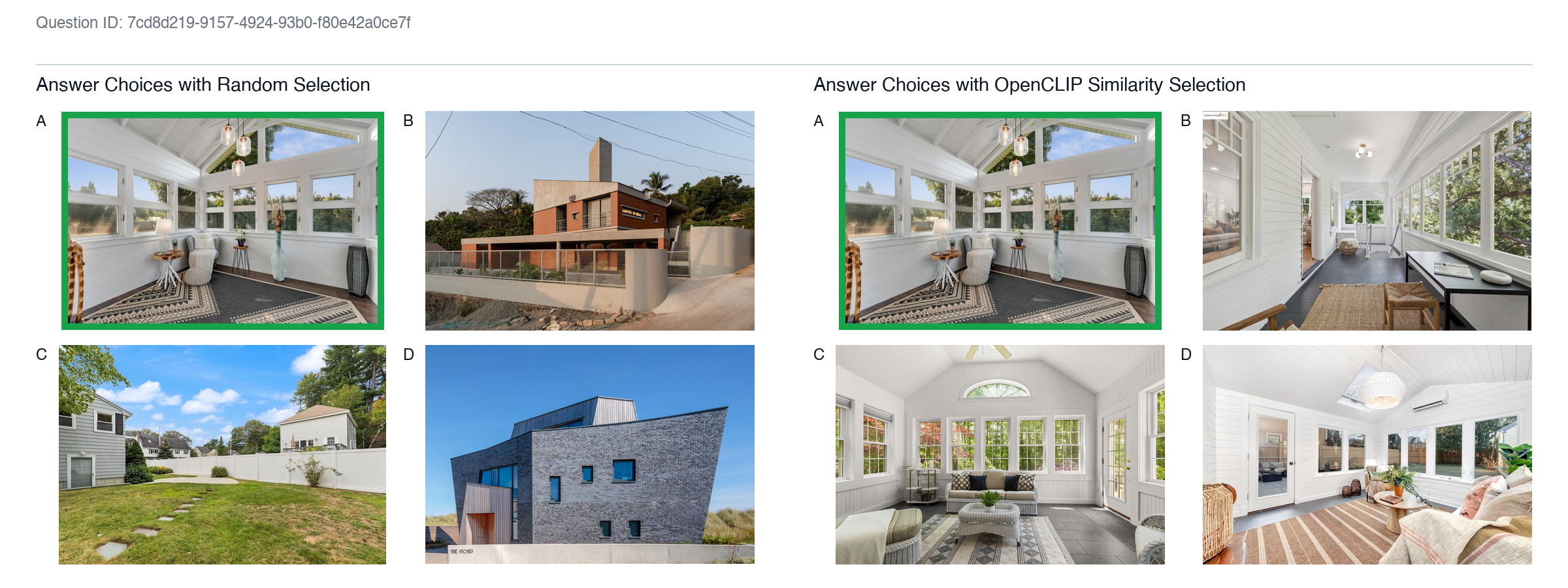}
  \caption{Random versus OpenCLIP similarity answer selection. Random selection produces visually unrelated distractors, whereas OpenCLIP similarity selection yields more plausible and visually similar answer choices, increasing the question’s difficulty. The question depicted is the same as in \cref{fig:example-question}.}
  \label{fig:similarity-selection}
\end{figure}

We used image embeddings to improve our image selection for incorrect answer choices. Once the correct answer image for a question sample was selected, we selected the 3 nearest images in the vector space based on the cosine similarity of the OpenCLIP image embeddings, filtering out any images that came from the same building record as the correct answer image. This procedure enabled us to select visually similar incorrect answer images, thereby increasing the difficulty of the resultant question samples (\cref{fig:similarity-selection}).

\subsection{LLM-in-the-loop Difficulty Selection}
\label{sec:difficulty-selection}

Automated question sample generation enabled us to generate a large number of candidate question samples, but not all generated question samples were suitable for inclusion in the final benchmark. We therefore required an automated method to filter the generated question sample set. Model-assisted filtering has previously been used to reduce benchmark shortcuts and retain more discriminative examples \citep{lebras2020aflite,yue2024mmmupro}. Agreement among the screening models provided a difficulty signal: items answered by all models were likely to be easy, whereas items missed by all models were potentially difficult, ambiguous, or invalid and therefore required manual review.

We generated 2,200 candidate question samples (200 for each archetype) and evaluated them against four state-of-the-art multimodal models: Claude Sonnet 4, Gemini 2.5 Pro, Pixtral Large, and OpenAI GPT-4.1. Questions answered correctly by at most half of the four screening models were treated as the primary pool for manual review. The model-difficulty-filtered pool contained 1,013 question samples at or below the 50\% threshold: 147 at 0\% model accuracy, 256 at 25\%, and 610 at 50\%. These 1,013 question samples were subsequently reviewed manually.

\subsection{Manual Curation for Final Benchmark Composition}
\label{sec:manual-curation}

The model-difficulty-filtered pool of 1,013 question samples was evaluated manually to select a final set of questions for the benchmark.

We built a dedicated validation interface for reviewing generated questions alongside aggregated model results. Reviewers knew the coarse screening bucket for the question (0\%, 25\%, or 50\% model accuracy), but they did not see model-specific answer choices or per-model correctness while making their initial judgment. Reviewers could first answer the question themselves and assess whether the item was solvable under the benchmark interface. They could then mark the question as valid or invalid.

Every question sample in the released benchmark was manually reviewed by two researchers. In practice, inclusion required that reviewers could identify a single correct answer, confirm that the images were of the correct type for the intended archetype, and determine that the item was human-solvable under the benchmark interface. Questions were rejected if they contained duplicate images, if the images were not of the correct type for the archetype, or if either reviewer judged the item unsolvable. When the two researchers initially disagreed, they reviewed the question together and reached consensus on whether to include it.

\subsection{Release Protocol}
\label{sec:release}

We have released ARCH-B as a JSON dataset containing the finalized question samples, answer choices, archetype metadata, and links to the publicly hosted images at their original source locations. For each image, we include a cryptographic hash of the exact source bytes used in our evaluation. We will not redistribute third-party image files, but will retain archival copies for integrity and preservation; researchers may contact the authors regarding unavailable assets, subject to applicable rights and institutional policies. The released question set will be preserved as a fixed benchmark version, with subsequent corrections or removals documented in a changelog and issued as new versions.

To support reproducible model evaluation, we have also released the benchmark runner code used to download images, serialize each JSON question, construct multimodal prompts, query models, parse responses, and score outputs. The runner will verify image hashes and flag unretrievable question sample images. Finally, we provide a link to the hosted human-study interface so that readers can interactively attempt representative questions and better understand the visual reasoning tasks evaluated by the benchmark (available at \href{https://archbench.codecolab.org/quiz}{https://archbench.codecolab.org/}). The released benchmark dataset and runner code accompany this preprint as arXiv ancillary files.

\section{Results}
\label{sec:results}

\subsection{Benchmark Composition}
\label{sec:composition}

The finalized benchmark contained 354 questions spanning 11 archetypes. \cref{app:archetypes} summarizes the final composition by archetype, including the final question count and dominant task for each archetype. The benchmark question samples were generated and curated according to the methods in \cref{sec:corpus,sec:generation,sec:difficulty-selection,sec:manual-curation}.

\subsection{Benchmark Evaluation Protocol}
\label{sec:evaluation-protocol}

The evaluation pipeline loaded each question from the benchmark JSON and converted it into a single ordered multimodal user message containing the question text, any reference images, and four answer choices labeled with unique IDs. Before transmission, images were downloaded, decoded, resized to a maximum dimension of 1,024 pixels while preserving aspect ratio, and re-encoded as JPEG at 90\% quality. All models received the answer choices in the same order and were instructed to return a structured output containing an answer choice id and a brief explanation.

Each model–question pair contributed exactly one scored outcome. Responses were processed using a deterministic parser, and an item was scored as correct only when the parsed answer ID exactly matched the ground-truth ID. The accompanying explanation was not used for scoring. If a model failed to return a response because of a timeout or terminal provider error, or returned an output that did not yield a syntactically valid answer ID, the outcome was counted as incorrect.

Across the 8,850 model-question-sample evaluations, 37 responses (0.42\%) ended in terminal inference failures and 559 responses (6.32\%) did not yield a valid answer ID. Per model valid response rates are reported alongside overall accuracy in \cref{tab:leaderboard}. Exact provider model identifiers, access dates, inference settings, and per-model failure counts are provided in the supplemental material. The complete evaluation runner, including prompt construction, image preprocessing, response parsing, and scoring, is released with the benchmark question set.

\subsection{Main Leaderboard}
\label{sec:leaderboard}

We evaluated 25 multimodal models on the full 354-question benchmark. \cref{tab:leaderboard} reports the exact leaderboard scores. Overall accuracy ranged from 10.45\% to 83.90\%, indicating that ARCH-B separates current systems across a wide performance band rather than saturating at either floor or ceiling. The strongest system was Gemini 3.1 Pro Preview, which answered 297 of 354 questions correctly (83.90\%). It was followed closely by Gemini 3.5 Flash (83.05\%), Claude Fable 5 (80.23\%), GPT-4.1 (77.97\%), and GPT-5.5 reasoning (76.55\%). Valid-response rates were above 94\% for most evaluated models, but substantially lower for Gemini 2.5 Flash (24.29\%), Claude Sonnet 4.6 (60.45\%), and Claude Sonnet 4.5 (68.08\%). Consequently, the low aggregate accuracies of these models partly reflect failure to produce an answer in the required format rather than incorrect selections among the four choices.

\begin{table}[t]
  \centering
  \scriptsize
  \setlength{\tabcolsep}{2pt}
  \begin{tabularx}{\columnwidth}{@{}r>{\raggedright\arraybackslash}Xl>{\raggedleft\arraybackslash}p{0.13\columnwidth}>{\raggedleft\arraybackslash}p{0.16\columnwidth}@{}}
    \toprule
    \textbf{Rank} & \textbf{Model} & \textbf{Family} & \textbf{Accuracy} & \textbf{Valid Response Rate} \\
    \midrule
    1 & gemini-3.1-pro-preview & Google & 83.90\% & 97.46\% \\
    2 & gemini-3.5-flash & Google & 83.05\% & 98.87\% \\
    3 & claude-fable-5 & Anthropic & 80.23\% & 98.87\% \\
    4 & gpt-4.1 & OpenAI & 77.97\% & 100.00\% \\
    5 & gpt-5.5 reasoning & OpenAI & 76.55\% & 99.72\% \\
    6 & gemini-3-flash-preview & Google & 75.99\% & 98.02\% \\
    7 & gpt-5.4 reasoning & OpenAI & 74.29\% & 99.72\% \\
    8 & gemini-2.5-pro & Google & 71.47\% & 100.00\% \\
    9 & gpt-5.5 & OpenAI & 70.90\% & 99.72\% \\
    10 & claude-opus-4-6-v1 & Anthropic & 70.34\% & 100.00\% \\
    11 & claude-opus-4-5-20251101-v1.0 & Anthropic & 65.25\% & 99.15\% \\
    12 & gpt-5.4 & OpenAI & 64.41\% & 98.87\% \\
    13 & grok-4.20-0309-reasoning & xAI & 57.06\% & 99.15\% \\
    14 & grok-4.3 & xAI & 50.28\% & 98.87\% \\
    15 & gpt-5.4-mini reasoning & OpenAI & 50.00\% & 100.00\% \\
    16 & gpt-5.4-mini & OpenAI & 46.61\% & 99.72\% \\
    17 & gemma3 27b & Google & 38.98\% & 99.72\% \\
    18 & claude-sonnet-4-6 & Anthropic & 37.85\% & 60.45\% \\
    19 & claude-sonnet-4-5-20250929-v1.0 & Anthropic & 34.46\% & 68.08\% \\
    20 & pixtral-large-2502-v1.0 & Mistral & 32.77\% & 97.18\% \\
    21 & gpt-5.4-nano reasoning & OpenAI & 29.10\% & 99.72\% \\
    22 & claude-haiku-4-5-20251001-v1.0 & Anthropic & 17.51\% & 94.63\% \\
    23 & gemini-2.5-flash & Google & 17.51\% & 24.29\% \\
    24 & gpt-5.4-nano & OpenAI & 16.95\% & 99.44\% \\
    25 & gemma3 4b & Ollama & 10.45\% & 100.00\% \\
    \bottomrule
  \end{tabularx}
  \caption{Main model leaderboard on ARCH-B.}
  \label{tab:leaderboard}
\end{table}


\subsection{Performance by Archetype}
\label{sec:archetype-performance}

\cref{tab:human-archetypes} summarizes model performance by archetype. Models are strongest on mixed-representation outlier detection, same-building photograph matching, view selection from mixed drawings, and elevation-to-photo matching, while their weakest categories include floorplan-to-photo matching and photograph-only outlier detection.

\subsection{Per-Question Difficulty Distribution}
\label{sec:question-difficulty}

\begin{figure}[t]
  \centering
  \includegraphics[width=\columnwidth]{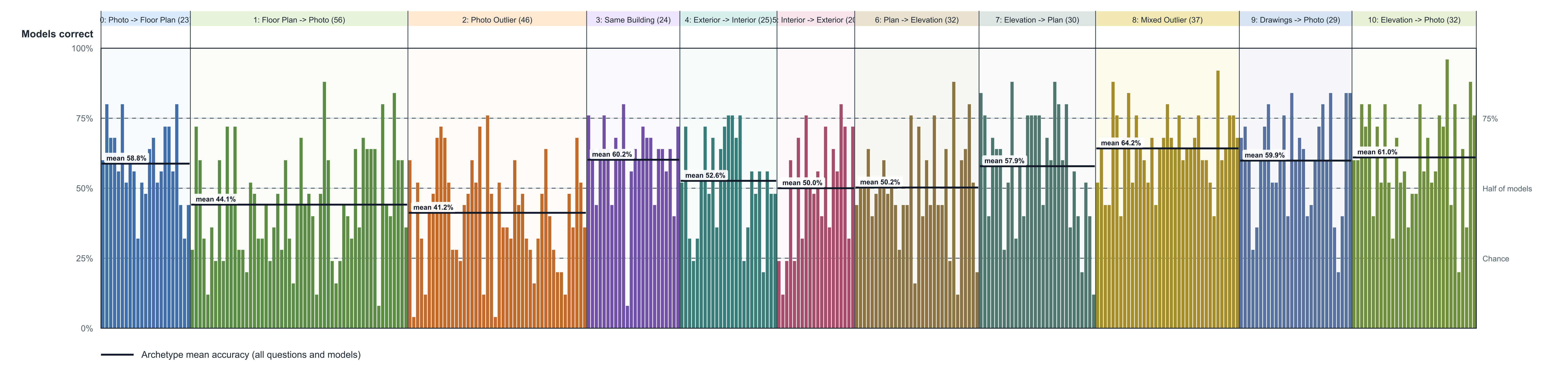}
  \caption{Per-question model accuracy across the 354-question benchmark. Each bar is one question, ordered and grouped by archetype; bar height gives the percentage of the 25 evaluated models that answered the question correctly.}
  \label{fig:question-accuracy}
\end{figure}

\cref{fig:question-accuracy} presents a per-question model-accuracy bar chart, showing that ARCH-B is not saturated for current multimodal models. Mean per-question model accuracy was 53.36\%, median accuracy was 56.00\%, and individual questions ranged from 4.00\% to 96.00\%. Crucially, there were no question samples answered correctly by every model (too easy) and no questions missed by every model (impossible). 145 of 354 questions were answered correctly by at most half of the evaluated models, 36 questions were at or below chance-level model accuracy, and 299 of 354 questions were answered correctly by at most 75\% of models rather than clustering near universal success. In combination with the manual validation described in \cref{sec:manual-curation}, this suggests that the final benchmark contains a graded spectrum of difficult but solvable cross-representational correspondence problems.


\subsection{Held-Out Model Analysis}
\label{sec:heldout}

To test whether LLM-in-the-loop selection introduced circularity, we grouped questions by how many of the four screening models answered them correctly and measured accuracy across all remaining models. Held-out accuracy increased monotonically from 28.65\% for questions answered by 0/4 screening models (\(n=33\)), to 41.38\% for 1/4 (\(n=77\)), and 59.02\% for 2/4 (\(n=244\)). Thus, the difficulty trends identified by the screening models generalized to the broader model set rather than reflecting model-specific weaknesses.


\section{Human Study}
\label{sec:human-study}

\subsection{Human Study Design}
\label{sec:human-design}

We conducted a human study to establish a human baseline score on the benchmark question sample set. Participants were recruited through Prolific, an online platform that connects researchers with verified study participants. The benchmark question sample set and evaluation interface were developed and hosted by us.

The study recruited English-speaking adults in the United States through the platform and forwarded them to our hosted evaluation interface. Each participant was presented with 10 question samples belonging to one archetype. The study was conducted in June 2026 and participants were paid \$2 per session. We recorded 583 completed sessions, resulting in 16 to 18 human responses per question sample. Our human study design was reviewed and approved by our institution’s IRB.

\subsection{Results}
\label{sec:human-results}

We collected 5,830 valid human responses across the 354 benchmark questions, with 16–18 responses per question. Participants answered 2,061 responses correctly, yielding a response-weighted accuracy of 35.35\% and a macro-average question accuracy of 35.37\%. Both measures exceed the four-choice chance level of 25\%. \cref{tab:human-archetypes} compares human accuracy with two summaries of model performance for each archetype: mean accuracy across all 25 evaluated models and the highest accuracy achieved by any evaluated model.

\begin{table*}[t]
  \centering
  \resizebox{\textwidth}{!}{%
  \begin{tabular}{@{}clrlrrr@{}}
    \toprule
    Archetype & Task family & Best acc. & Best Model for Archetype & Mean model acc. & Human acc. & Best-human gap \\
    \midrule
    \textbf{0} & Match floor plan with images & 95.65 & gemini-3.1-pro-preview & 58.78 & 34.74 & 60.91 \\
    \textbf{1} & Match image with floor plan & 76.79 & gpt-4.1 & 44.14 & 36.41 & 40.38 \\
    \textbf{2} & Choose outlying image & 84.78 & gemini-3.1-pro-preview & 41.22 & 21.61 & 63.17 \\
    \textbf{3} & Choose inlying image & 100.00 & gpt-4.1 & 60.17 & 32.00 & 68.00 \\
    \textbf{4} & Identify interior from exterior & 96.00 & gemini-3.1-pro-preview & 52.64 & 35.85 & 60.15 \\
    \textbf{5} & Identify exterior from interior & 85.00 & claude-opus-4-5-20251101-v1.0 & 50.00 & 29.70 & 55.30 \\
    \textbf{6} & Identify elevation from floor plan & 90.62 & gemini-3.5-flash & 50.25 & 36.95 & 53.67 \\
    \textbf{7} & Identify floor plan from elevation & 90.00 & gemini-3.5-flash & 57.87 & 32.25 & 57.75 \\
    \textbf{8} & Identify outlier from mixed representations & 100.00 & claude-fable-5 & 64.22 & 33.93 & 66.07 \\
    \textbf{9} & Identify view from mixed drawings & 96.55 & gemini-3-flash-preview & 59.86 & 42.44 & 54.11 \\
    \textbf{10} & Identify view from elevations & 96.88 & gemini-3.5-flash & 61.00 & 56.15 & 40.73 \\
    \bottomrule
  \end{tabular}
  }
  \caption{Archetype-level accuracy.}
  \label{tab:human-archetypes}
\end{table*}

The mean-model and best-observed results describe different aspects of model performance. Mean model accuracy indicates how difficult an archetype was across the evaluated model set. Best observed accuracy reveals state of the art performance per archetype. Gemini 3.1 Pro Preview produced the best result for four archetypes, Gemini 3.5 Flash for three, GPT-4.1 for two, and Claude Fable 5, Claude Opus 4.5, and Gemini 3 Flash Preview for one each.

Human accuracy was highest for elevation-to-photograph matching (56.15\%) and selecting a view from mixed drawings (42.44\%), and lowest for photograph-only outlier detection (21.61\%) and exterior-from-interior matching (29.70\%). Mean model accuracy was highest for mixed-representation outlier detection (64.22\%) and lowest for photograph-only outlier detection (41.22\%) and floorplan-to-photograph matching (44.14\%). The best observed model accuracy ranged from 76.79\% to 100\%, exceeding human accuracy in every archetype by between 40.38 and 68.00 percentage points.

The relative ordering of archetypes differs between humans and models. Human accuracy and mean model accuracy have a weak rank correlation across archetypes (Spearman’s \(\rho=0.33\)), indicating limited agreement in which tasks are comparatively easy or difficult. \cref{sec:failure-modes} considers the possible failure modes underlying these differences.

\section{Discussion}
\label{sec:discussion}

\subsection{Benchmark Saturation and Continued Use}
\label{sec:saturation}

Although the strongest evaluated model scores above 80\%, ARCH-B still provides value as a comparative benchmark because performance is far from uniform across models and task families. Overall leaderboard accuracy spans a wide range, from 10.45\% to 83.90\%, and the archetype-level results show that models differ not only in aggregate strength but in the kinds of architectural comprehension they handle well. This makes the benchmark useful as a way to measure relative capability: whether a model is strong on mixed-representation outlier detection, plan-to-elevation correspondence, interior-exterior matching, or floorplan-to-photograph transfer. In this sense, the benchmark's value comes from its breadth and diagnostic structure. A high top-line score does not eliminate the need to understand which representational transitions are easy, which remain brittle, and whether improvements in one task family generalize to others.

The question generation pipeline also has value beyond leaderboard evaluation; the same pipeline could be used to generate larger quantities of annotated data for targeted training or fine-tuning. For example, if a model is weak on floorplan-to-photograph matching or elevation-to-plan correspondence, the system could generate additional supervised examples for that specific task family rather than treating architectural visual understanding as a single undifferentiated capability.

\subsection{Human and Model Failure Modes}
\label{sec:failure-modes}

The archetype-level results indicate that humans and models have overlapping but non-identical patterns of difficulty.

Photograph-only outlier detection was the weakest archetype for both groups, with 21.61\% human accuracy and 41.22\% mean model accuracy. Unlike positive matching, this task requires establishing consistency among several views while rejecting a visually similar photograph; agreement in materials, style, or context may therefore be mistaken for building identity. Interior-to-exterior matching was also difficult, particularly for humans (29.70\%), plausibly because an interior provides only local and often non-distinctive evidence about the building's exterior form.

 Models exhibited a directional asymmetry that was largely absent for humans: mean model accuracy was 58.78\% when selecting a floor plan from photographs, but only 44.14\% when selecting a photograph from floor plans, whereas human accuracy was similar in the two directions (34.74\% and 36.41\%). This suggests that models can use the stable topology of candidate plans to explain observed views more readily than they can project an abstract plan into a plausible visual appearance. Conversely, humans performed best when matching elevations to photographs (56.15\%), where facade proportions, window rhythms, and massing provide comparatively direct visual correspondences. The model advantage on mixed-representation outlier detection (64.22\% versus 33.93\%) may reflect a greater ability to pool weak cues across several representations, while the lower non-expert human result may partly reflect unfamiliarity with architectural drawings. These directional asymmetries are consistent with prior evidence that multimodal performance on visual correspondence depends strongly on viewpoint and representational form \citep{fu2024blink,yang2025thinking,ma2025threedsrbench}.

\subsection{Human Study Improvements and Future Work}
\label{sec:future-work}

The human study is best interpreted as a non-expert baseline for the benchmark rather than a general comparison between people and models under all scenarios. Participants completed short sessions and were paid for completion rather than accuracy, so low performance may reflect architectural inexperience, limited attention, or unfamiliarity with plans and elevations in addition to item difficulty; related evaluations likewise find differences associated with architectural training \citep{shen2026archsibench}. Future studies should compare non-experts, architecture students, and practitioners, record response time and confidence, and test whether incentives or brief instruction improve performance.

Future versions of ARCH-B could expand beyond photographs, plans, elevations, sections, and renderings to include axonometric drawings, diagrams, site plans, construction documents, three-dimensional models, and images from different stages of construction or use. They could also move beyond fixed answer choices to require localization, spatial ordering, evidence-based explanations, or reconstruction of relationships among rooms, floors, facades, and buildings. Controlled variation in distractor similarity, drawing style, image quality, building type, and geographic source would help distinguish robust representation transfer from reliance on superficial cues.

\section{Conclusion}
\label{sec:conclusion}

ARCH-B evaluates whether multimodal models can align photographs, plans, elevations, sections, and mixed evidence as representations of the same building. Evaluation across 25 models and a non-expert human baseline reveals substantial differences among systems and uneven performance across task types and individual questions. ARCH-B is therefore not a general measure of architectural intelligence, but a reusable stress test for visual correspondence and representation transfer across architectural media.

\section*{Acknowledgments}

The research in this paper was funded by a grant from the Joint Center for Housing Studies of Harvard University. Earlier stages of this research, and technical infrastructural support were funded by the Daniel Schodek Fund for Research in Design Technology at Harvard University. We thank Martin Bechthold for his support, guidance and feedback throughout the project.

The authors used AI tools in all stages of this research, including, but not restricted to, for literature review, tools development and editing of the manuscript. The authors reviewed and edited all outputs and take full responsibility for the content of this publication.

\appendix
\section{Question Archetypes}
\label{app:archetypes}

\begingroup
\setlength{\parindent}{0pt}
\newcommand{\archetypeentry}[7]{%
  \noindent\begin{minipage}[t]{\columnwidth}
    \centering
    \includegraphics[width=0.85\columnwidth,trim=0 0 0 165,clip]{#7}\\[-0.35ex]
    {\fontsize{5}{5.35}\selectfont\raggedright
      \textbf{#1. #2}\hfill\textbf{\textit{n}:} #6\par
      \textbf{Question:} #3\par
      \textbf{Refs:} #4\quad\textbf{Choices:} #5\par}
  \end{minipage}\par\vspace{0.35ex}%
}
\archetypeentry{0}{Match floor plan with images}{Choose the floorplan that corresponds to the building shown in these photographs.}{2 to 4 Photographs}{4 Floor Plans}{23}{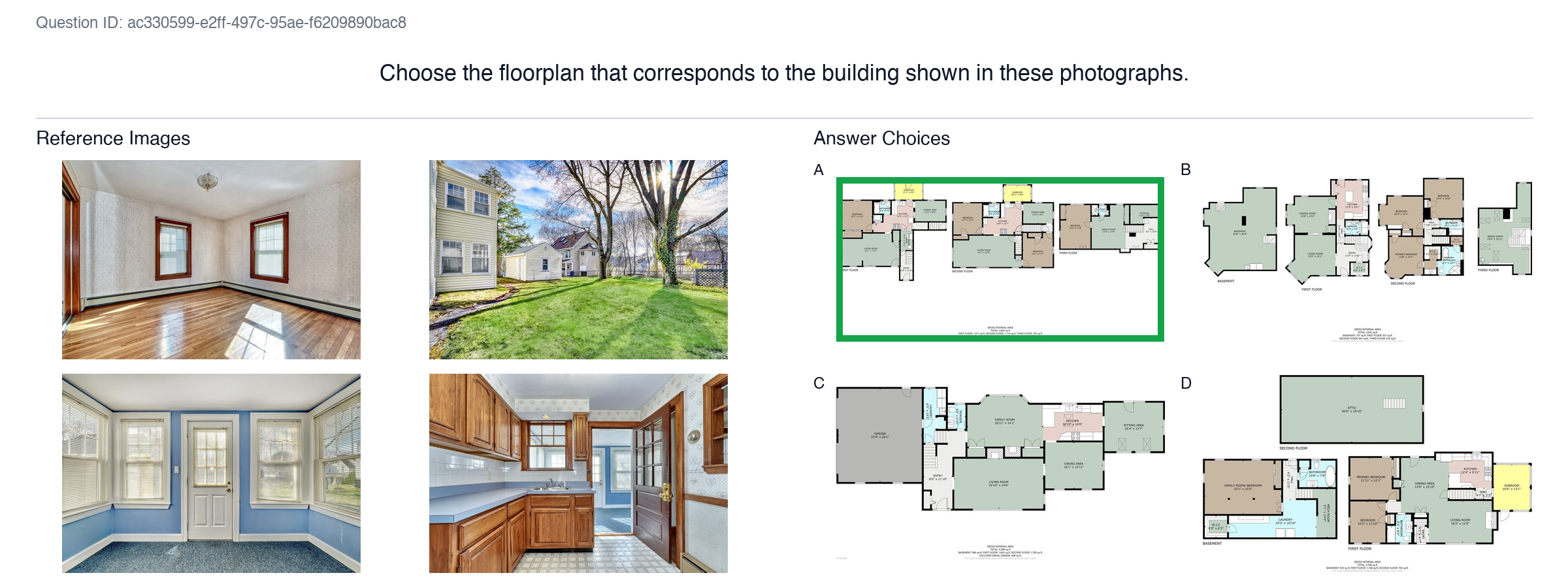}
\archetypeentry{1}{Match image with floor plan}{Choose the photograph that corresponds to the building shown in this/these floorplan/s.}{1 to 4 Floor Plans}{4 Photographs}{56}{figures/archetype_02_choose-the-photograph-that-corresponds-to-the-building_7cd8d219.png}
\archetypeentry{2}{Choose outlying image}{One of these photographs does NOT belong to the same building as the other three. Choose the OUTLIER.}{None}{4 Photographs}{46}{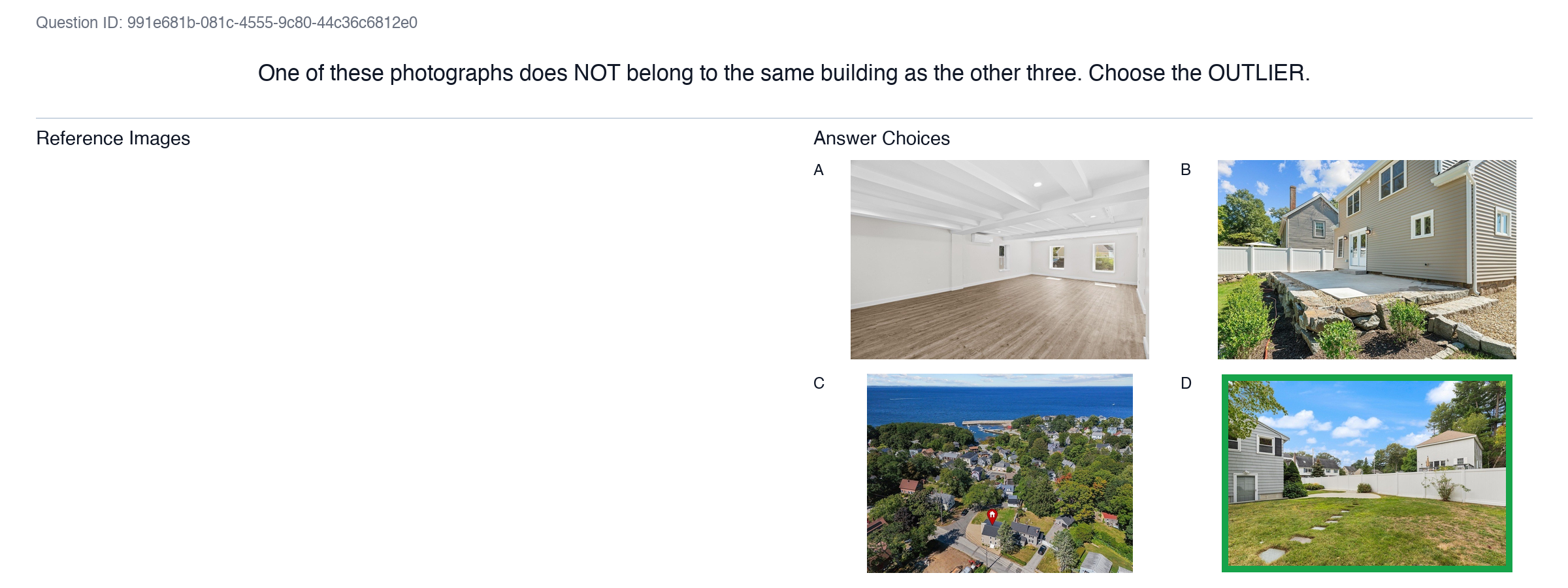}
\archetypeentry{3}{Choose inlying image}{Choose the photograph that corresponds to the same building as these three.}{3 Photographs}{4 Photographs}{24}{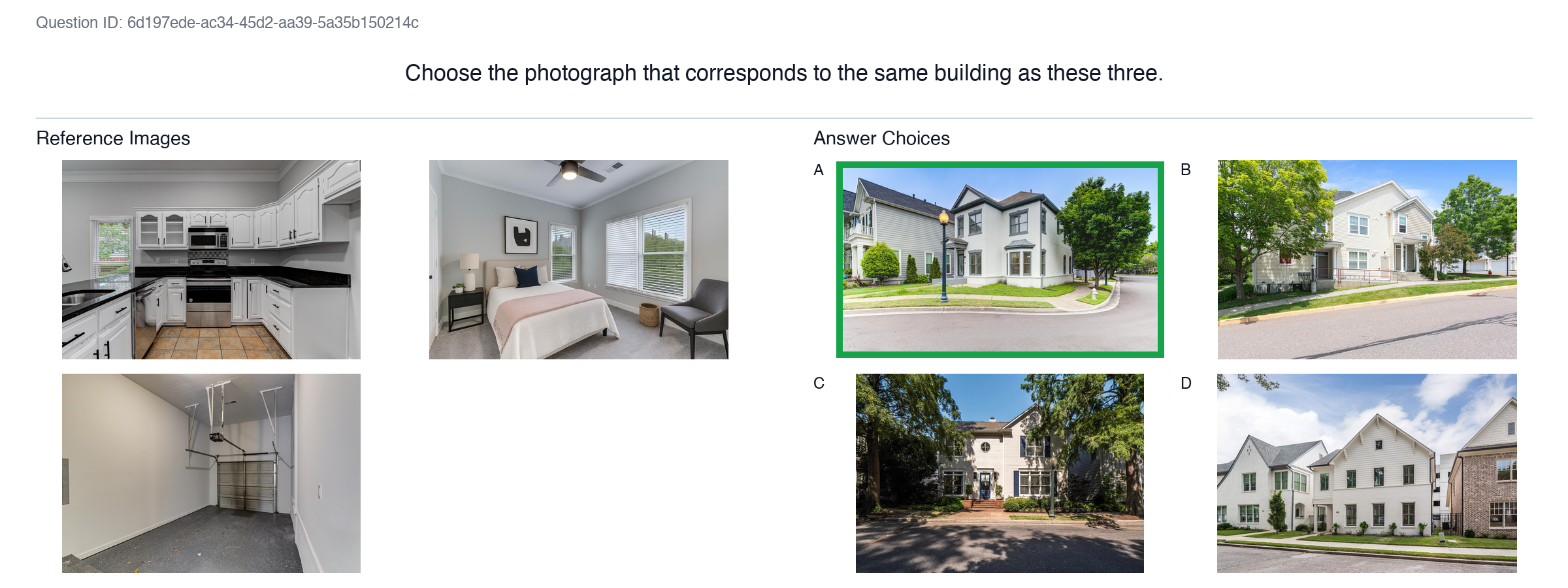}
\archetypeentry{4}{Identify interior from exterior}{Choose the interior photograph that corresponds to the same building as these exterior ones.}{4 Exterior Photographs}{4 Interior Photographs}{25}{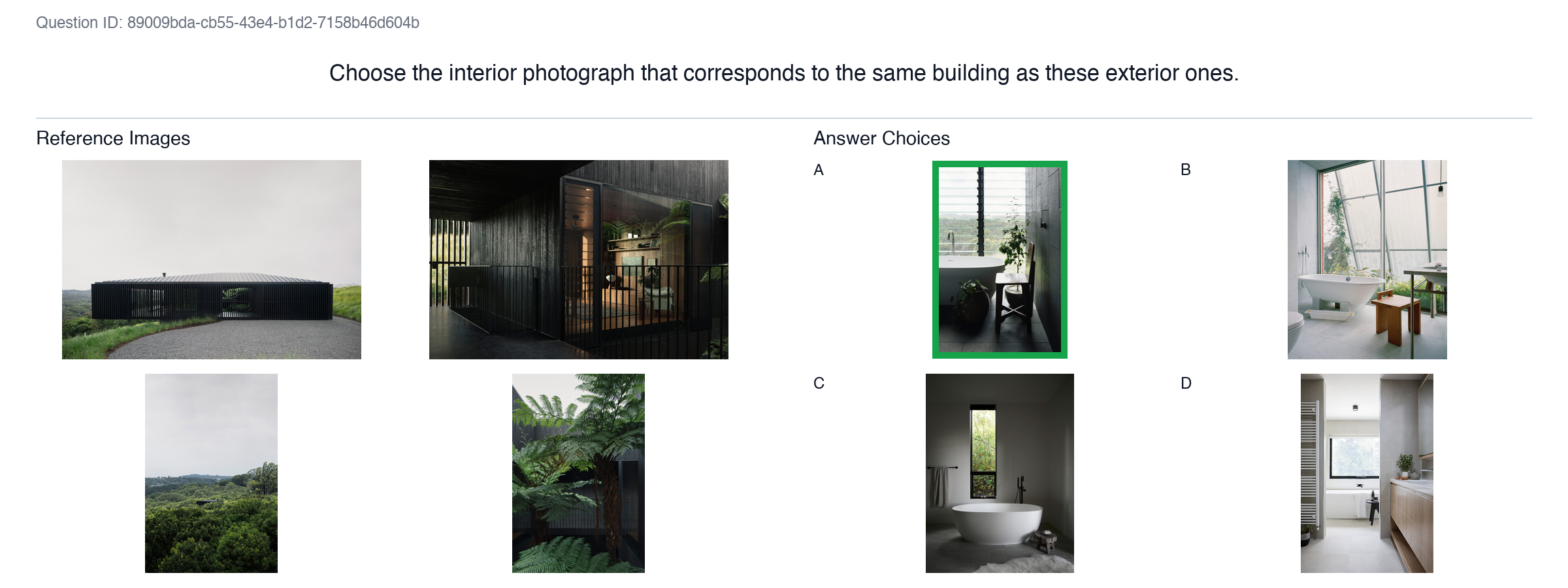}
\archetypeentry{5}{Identify exterior from interior}{Choose the exterior photograph that corresponds to the same building as these interior ones.}{4 Interior Photographs}{4 Exterior Photographs}{20}{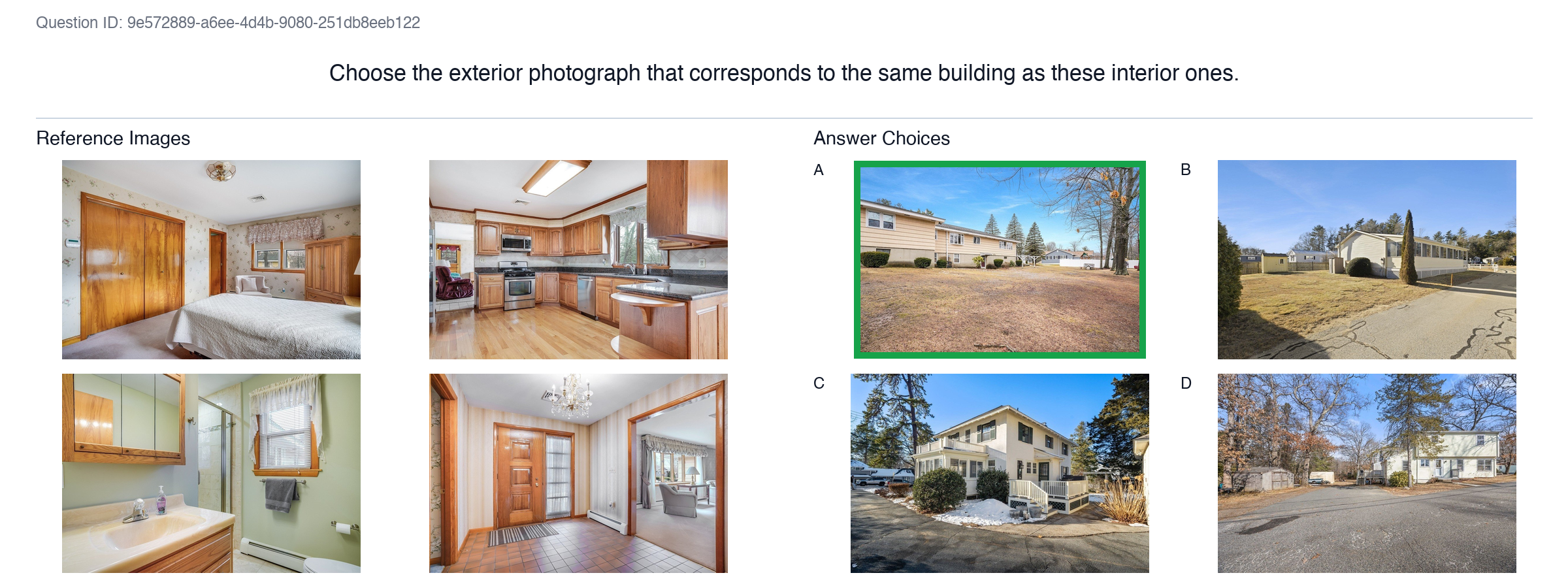}
\archetypeentry{6}{Identify elevation from floor plan}{Choose the elevation drawing that corresponds to the same building as this/these plan/s.}{1 to 4 Floor Plans}{4 Elevation Drawings}{32}{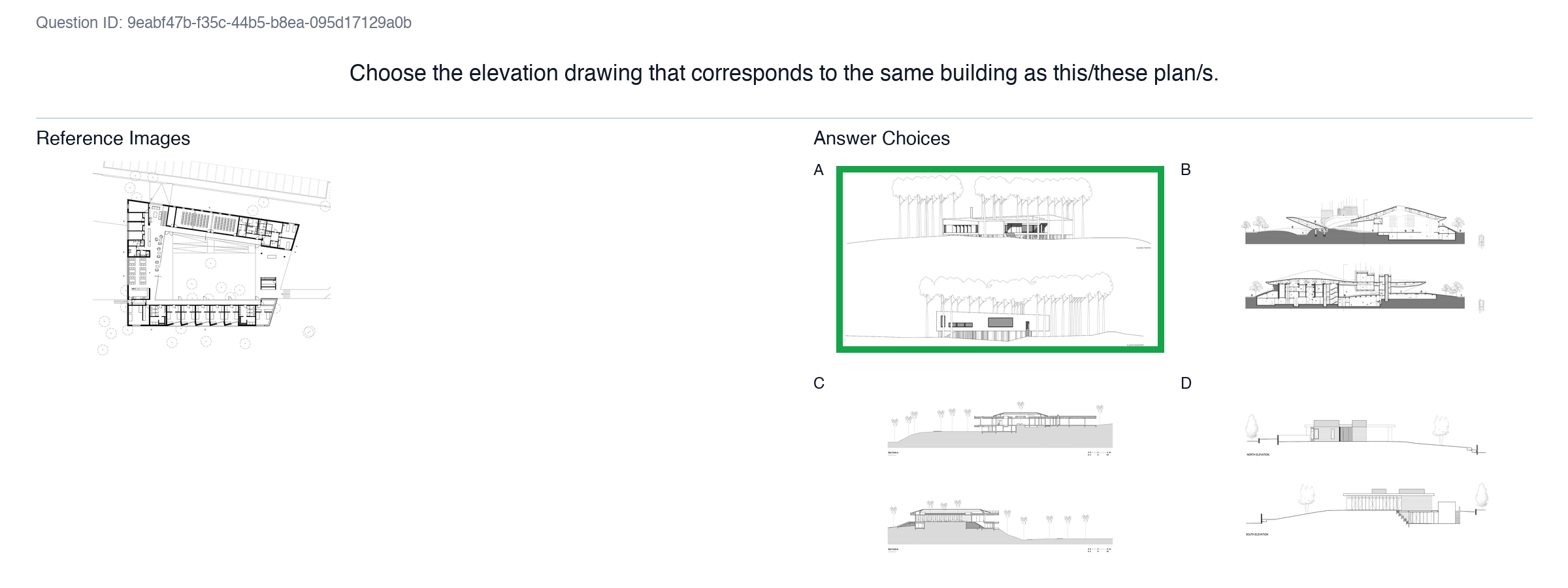}
\archetypeentry{7}{Identify floor plan from elevation}{Choose the plan drawing that corresponds to the same building as this/these elevation/s.}{1 to 4 Elevation Drawings}{4 Floor Plans}{30}{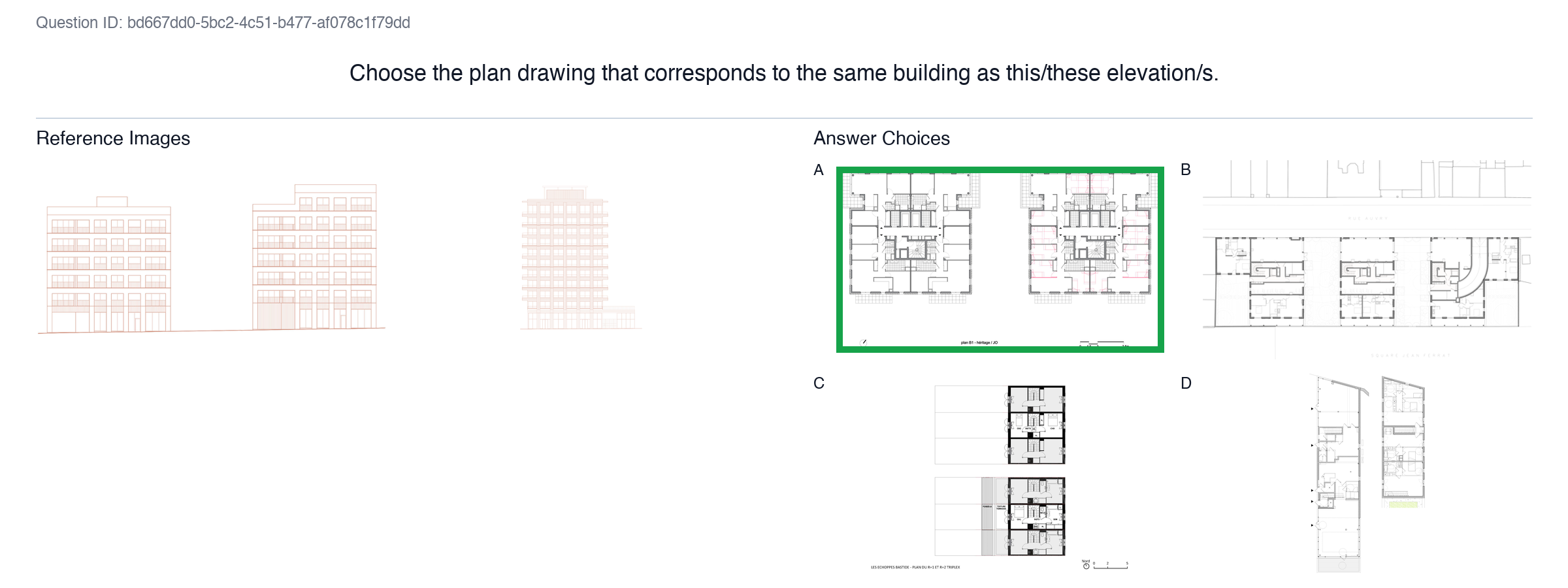}
\archetypeentry{8}{Identify outlier from mixed representations}{One of these photographs or drawings does NOT belong to the same building as the other three. Choose the outlier.}{None}{1 Floor Plan, 1 Elevation Drawing, 1 Section Drawing, and 1 Photograph}{37}{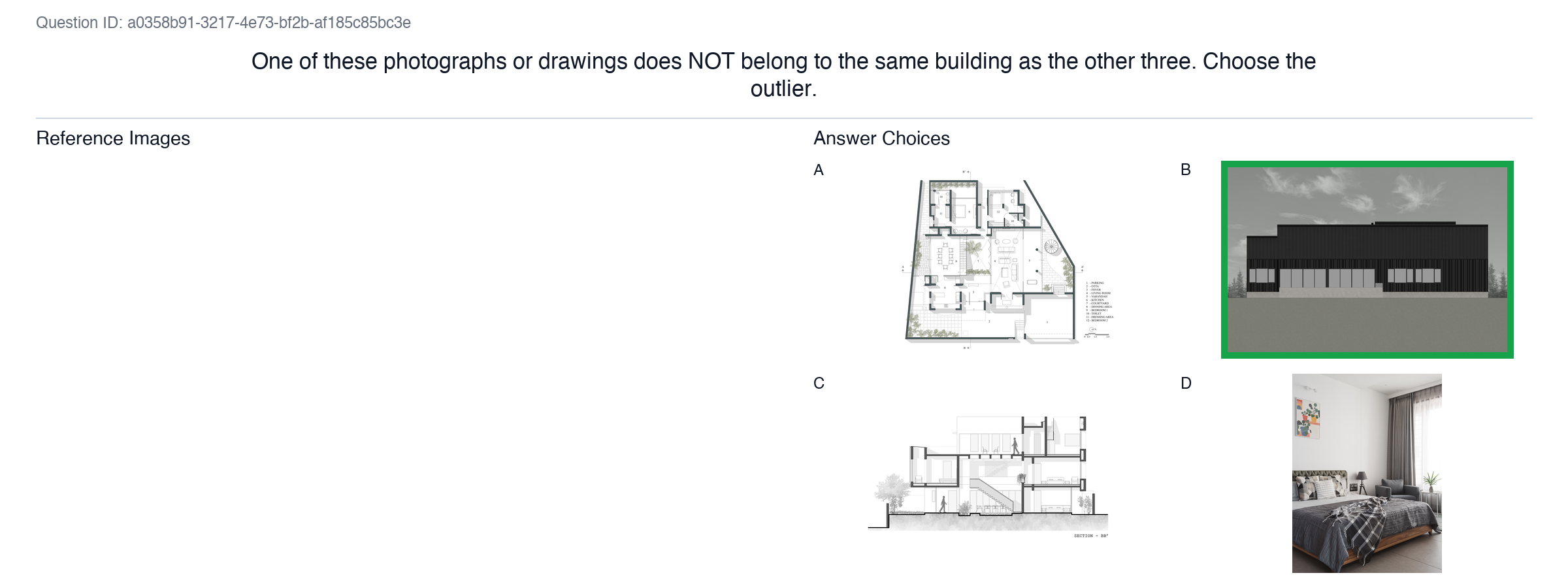}
\archetypeentry{9}{Identify view from mixed drawings}{Choose the photograph that corresponds to the building shown in these drawings.}{1 Floor Plan, 1 Elevation Drawing, and 1 Section Drawing}{4 Photographs}{29}{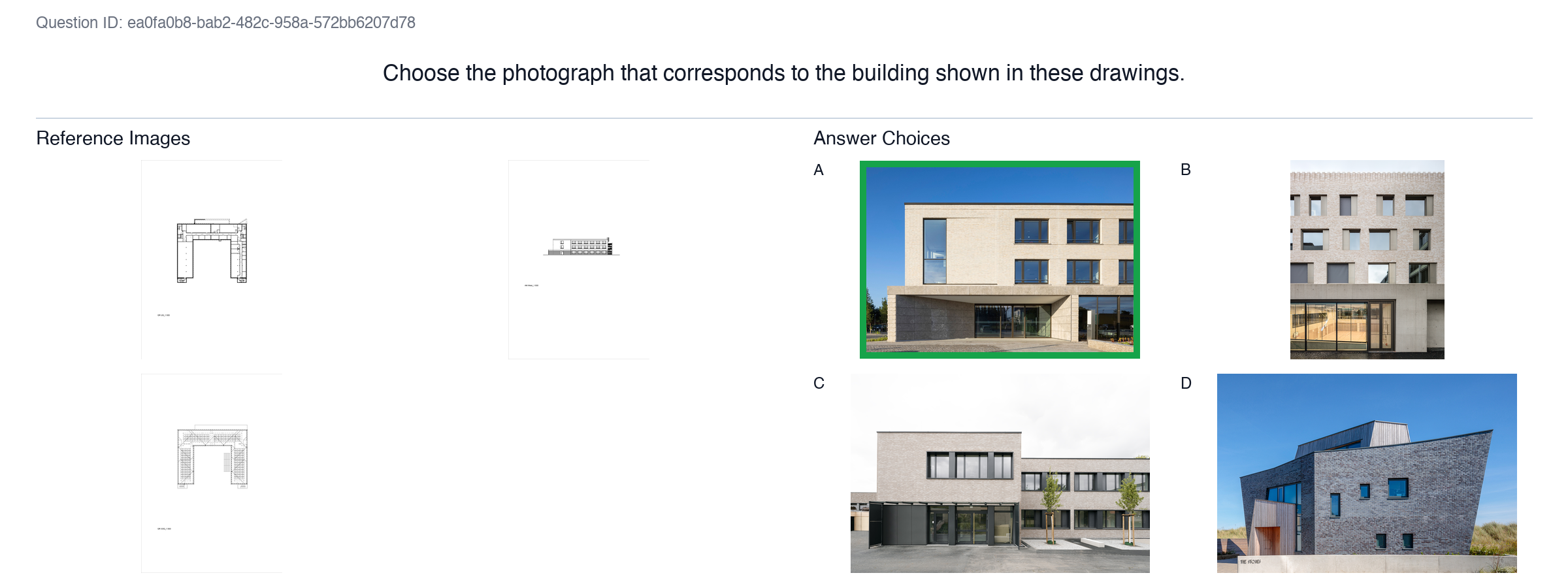}
\archetypeentry{10}{Identify view from elevations}{Choose the photograph that corresponds to the building shown in these elevations.}{2 Elevation Drawings}{4 Exterior Photographs}{32}{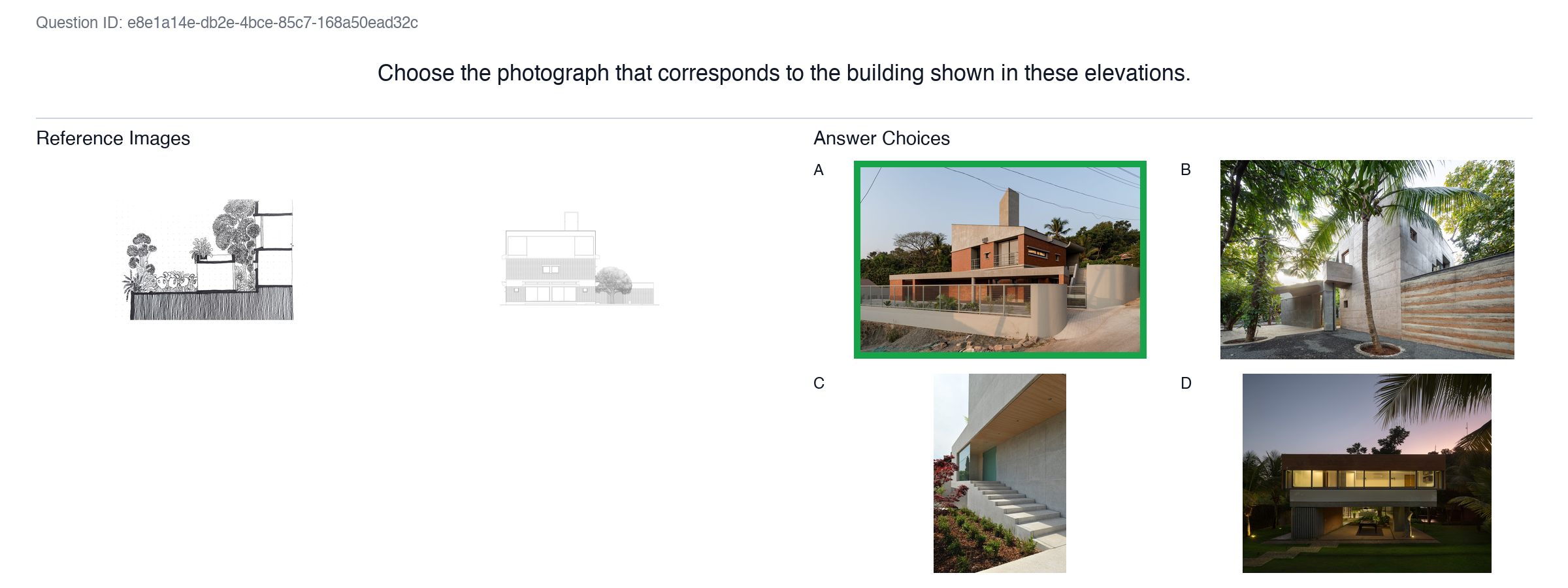}
\endgroup

{
  \small
  \bibliographystyle{ieeenat_fullname}
  \bibliography{main}
}

\end{document}